\documentclass{article} 
\usepackage{iclr2021_conference,times}
\usepackage{xcolor}
\usepackage{textcomp}
\usepackage{hyperref}
\hypersetup{
 colorlinks=True,
 linkcolor=blue,
 citecolor=blue,
 urlcolor=blue}

\usepackage{amsmath,amsfonts,bm}
\usepackage{microtype}
\usepackage{graphicx}
\usepackage{booktabs} 
\usepackage{times}
\usepackage{epsfig}
\usepackage{graphicx}
\usepackage{amssymb}
\usepackage{subfig}
\usepackage{enumitem}
\usepackage{wrapfig}
\usepackage[ruled]{algorithm2e}

\def\eqref#1{equation~\ref{#1}}

\def\1{\bm{1}}

\DeclareMathAlphabet{\mathsfit}{\encodingdefault}{\sfdefault}{m}{sl}
\SetMathAlphabet{\mathsfit}{bold}{\encodingdefault}{\sfdefault}{bx}{n}

\newcommand{\E}{\mathbb{E}}

\usepackage{hyperref}
\usepackage{wrapfig}
\usepackage{listings}
\usepackage{url}

\newcommand{\ar}[1] {{\color{magenta} AR: {#1}}}
\newcommand{\wendy}[1] {{\color{blue} Wendy: {#1}}}

\definecolor{codeblue}{rgb}{0.25,0.5,0.5}
\definecolor{darklavender}{rgb}{0.45, 0.31, 0.59}
\definecolor{htmlgreen}{rgb}{0.0, 0.5, 0.0}

\definecolor{keywords}{rgb}{0.08,0.54,.02}
\newcommand\blfootnote[1]{%
  \begingroup
  \renewcommand\thefootnote{}\footnote{#1}%
  \addtocounter{footnote}{-1}%
  \endgroup
}

\title{Reinforcement Learning with Latent Flow}

\author{Wenling Shang$^{2,\dagger}$ , Xiaofei Wang$^{1,\dagger}$, \\
\textbf{Aravind Srinivas$^1$, Aravind Rajeswaran$^3$, Yang Gao$^1$}, \\
\textbf{Pieter Abbeel$^1$ \& Michael Laskin$^1$
} \\
University of California Berkeley$^1$, Deepmind$^2$, University of Washington$^3$\\
}

\iclrfinalcopy 
\begin{document}

\maketitle
\blfootnote{$\dagger$: equal contribution. Correspondence: \href{mailto:wendyshang@google.com}{wendyshang@google.com}. Code: \url{https://github.com/WendyShang/flare}. }
\vspace*{-10pt}

\begin{abstract}

Temporal information is  essential to learning effective policies with Reinforcement Learning (RL). However, current state-of-the-art RL algorithms either assume that such information is given as part of the state space or, when learning from pixels, use the simple heuristic of frame-stacking to implicitly capture temporal information present in the image observations. This heuristic is in contrast to the current paradigm in video classification architectures, which utilize explicit encodings of temporal information through methods such as optical flow and two-stream architectures to achieve state-of-the-art performance. Inspired by leading video classification architectures, we introduce the {\bf F}low of {\bf La}tents for {\bf Re}inforcement Learning (\emph{Flare}), a network architecture for RL that explicitly encodes temporal information through latent vector differences. We show that Flare (i) recovers optimal performance in state-based RL without explicit access to the state velocity, solely with positional state information, (ii) achieves state-of-the-art performance on pixel-based challenging continuous control tasks within the DeepMind control benchmark suite, namely quadruped walk, hopper hop, finger turn hard, pendulum swing, and walker run, and is the most sample efficient model-free pixel-based RL algorithm, outperforming the prior model-free state-of-the-art by $\textbf{1.9}\times$ and $\textbf{1.5}\times$ on the 500k and 1M step benchmarks, respectively, and (iv), when augmented over rainbow DQN, outperforms this state-of-the-art level baseline on 5 of 8 challenging Atari games at 100M time step benchmark.

\end{abstract}

\section{Introduction}
Reinforcement learning (RL)~\citep{SuttonBook} holds the promise of enabling artificial agents to solve a diverse set of tasks in uncertain and unstructured environments. Recent developments in RL with deep neural networks have led to tremendous advances in autonomous decision making. Notable examples include classical board games~\citep{alphago, alphagozero}, video games~\citep{mnih2015human, openai2019dota, alphastar}, and continuous control~\citep{ppo, ddpg, DAPG}. A large body of research has focused on the case where an RL agent is equipped with a compact state representation. 
%
Such compact state representations are typically available in simulation~\citep{mujoco, tassa2018deepmind} or in laboratories equipped with elaborate motion capture systems~\citep{OpenAIHand, Zhu2019DexterousMW, Lowrey2018ReinforcementLF}. However, state representations are seldom available in unstructured real-world settings like the home. For RL agents to be truly autonomous and widely applicable, sample efficiency and the ability to act using raw sensory observations like pixels is crucial. Motivated by this understanding, we study the problem of efficient and effective deep RL from pixels.

A number of recent works have made progress towards closing the sample-efficiency and performance gap between deep RL from states and pixels~\citep{laskin_srinivas2020curl,laskin_lee2020rad,hafner2019dream,kostrikov2020image}. An important component in this endeavor has been the extraction of high quality visual features during the RL process. \citet{laskin_lee2020rad} and \citet{stooke2020atc} have shown that features learned either explicitly with auxiliary losses (reconstruction or contrastive losses) or implicitly (through data augmentation) are sufficiently informative to recover the agent's pose information. While existing methods can encode positional information from images, there has been little attention devoted to extracting temporal information from a stream of images. As a result, existing deep RL methods from pixels struggle to learn effective policies on more challenging continuous control environments that deal with partial observability, sparse rewards, or those that require precise manipulation.

 \begin{wrapfigure}{r}{0.5\textwidth}
     \centering
     \vspace{-5mm}
     \includegraphics[width=0.48\textwidth]{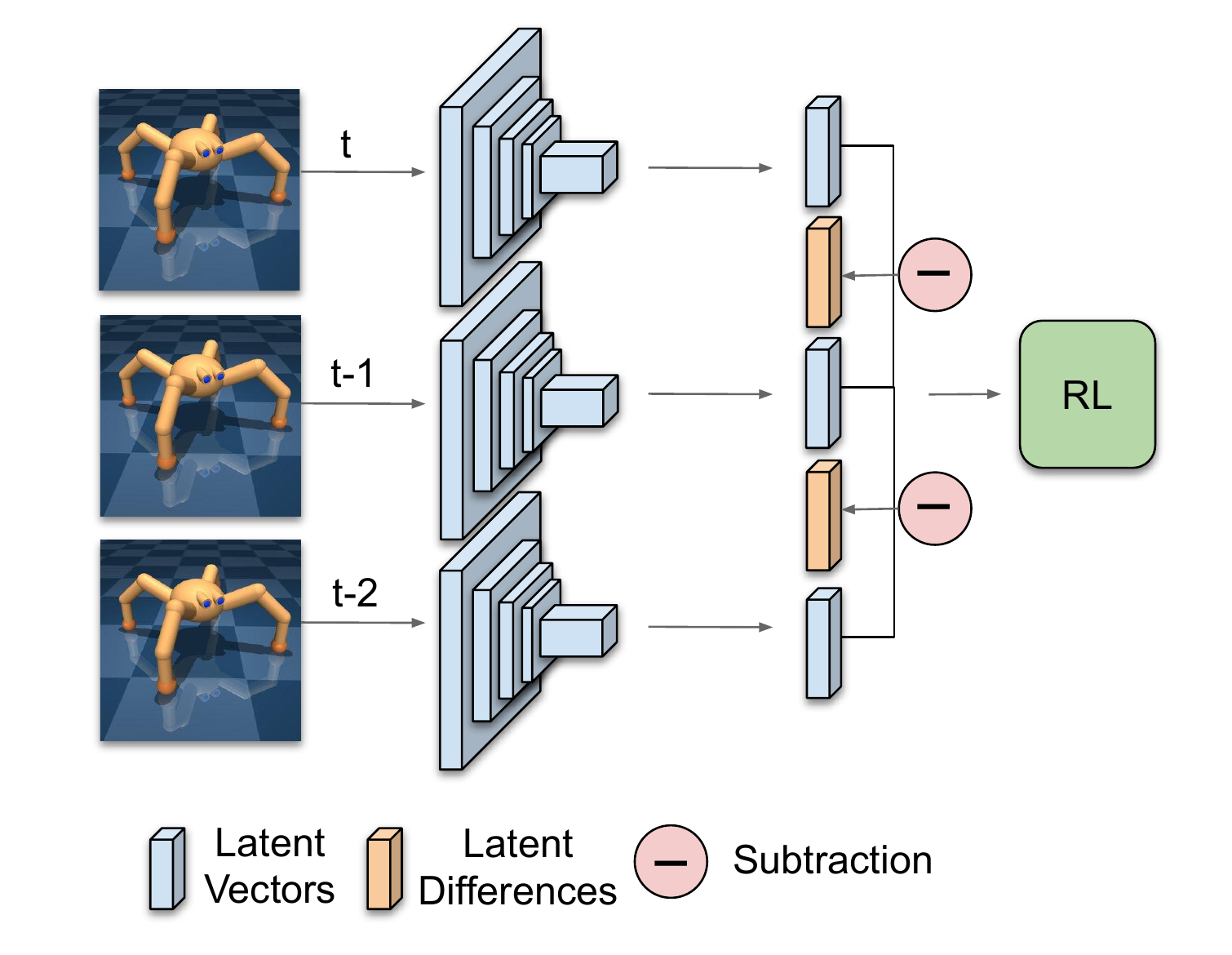}
     \vspace{-2mm}
     \small
     \caption{{\bf F}low of {\bf La}tents for {\bf Re}inforcement Learning (Flare) architecture. Input frames are first encoded individually by the same encoder. The resulting latent vectors are then concatenated with their latent differences before being passed to the downstream RL algorithm. }
    \vspace{-5mm}
     \label{fig:flare_intro}
 \end{wrapfigure}
 
Current approaches in deep RL for learning temporal features are largely heuristic in nature. A commonly employed approach is to stack the most recent frames~\citep{mnih2015human} as inputs to a convolutional neural network (CNN). This can be interpreted as a form of early fusion~\citep{KarpathyVideo}, where information from the recent time window is combined immediately at the pixel level for input to the CNN. In contrast, modern video recognition systems use alternate architectures that employ optical flow and late fusion~\citep{simonyan2014two}, where frames are processed individually with CNN layers before fusion and downstream processing. Such a late fusion approach is typically beneficial due to better performance, fewer parameters, and the ability to use multi-modal data~\citep{ Chebotar2017PathIG, Jain-ICRA-19}. However, directly extending such architectures to RL is be challenging.
Real-time computation of optical flow for action selection can be computationally infeasible for many applications with fast control loops like robotics. Furthermore, optical flow computation at training time can also be prohibitively expensive. 
In our experiments, we also find that a naive late fusion architecture minus the optical flow yields poor results in RL settings (see Section~\ref{sec:ablate}). This observation is consistent with recent findings in related domains like visual navigation~\citep{Walsman-IROS-19}.

To overcome the above challenges, we develop {\bf F}low of {\bf La}tents for {\bf Re}inforcement Learning (\emph{Flare}), a new architecture for deep RL from pixels (Figure~\ref{fig:flare_intro}). Flare can be interpreted as a {\em structured late fusion} architecture. Flare processes each frame individually to compute latent vectors, similar to a standard late fusion approach (see Figure~\ref{fig:flare_intro}). Subsequently, temporal differences between the latent feature vectors are computed and fused along with the latent vectors by concatenation for downstream processing. By incorporating this structure of temporal difference in latent feature space, we provide the learning agent with appropriate inductive bias. In experiments, we show that Flare (i) recovers optimal performance in state-based RL without explicit access to the state velocity, solely with positional state information, (ii) achieves state-of-the-art performance compared to model-free methods on several challenging pixel-based continuous control tasks within the DeepMind control benchmark suite, namely Quadruped Walk, Hopper Hop, Finger Turn-hard, Pendulum Swingup, and Walker Run, while being the most sample efficient model-free pixel-based RL algorithm across these tasks, outperforming the prior model-free state-of-the-art RAD by $\textbf{1.9}\times$ and $\textbf{1.5}\times$ on the 500k and 1M environment step benchmarks, respectively, and (iii) when augmented over Rainbow DQN, outperforms the baseline on 5 out of 8 challenging Atari games at 100M step benchmark.

\section{Related Work}
\textbf{Pixel-Based RL} The ability of an agent to autonomously learn control policies from visual inputs can greatly expand the applicability of deep RL~\citep{dosovitskiy2017carla, savva2019habitat}. Prior works have used CNNs to extend RL algorithms like PPO~\citep{ppo}, SAC~\citep{haarnoja2018soft}, and Rainbow~\citep{hessel2017rainbow} to pixel-based tasks. Such direct extensions have typically required substantially larger number of environment interactions when compared to the state-based environments. In order to improve sample efficiency, recent efforts have studied the use of auxiliary tasks and loss functions~\citep{yarats2019improving, laskin_srinivas2020curl, schwarzer2020data}, data augmentation~\citep{laskin_lee2020rad, kostrikov2020image}, and latent space dynamics modeling~\citep{hafner2019learning, hafner2019dream}. Despite these advances, there is still a large gap between the learning efficiency in state-based and pixel-based environments in a number of challenging benchmark tasks. Our goal in this work is to identify where and how to improve pixel-based performance on this set of challenging control environments. 

\textbf{Neural Network Architectures in RL} The work of \citet{mnih2015human} combined Q-learning with CNNs to achieve human level performance in Atari games, wherein \citet{mnih2015human} concatenate the most recent $4$ frames and use a convolutional neural network to output the Q values. In 2016, \citet{mnih2016asynchronous} proposed to use a shared CNN among frames to extract visual features and aggregate the temporal information with LSTM. The same architectures have been adopted by most works to date~\citep{laskin_srinivas2020curl, schwarzer2020data,kostrikov2020image, laskin_lee2020rad}. 
The development of new architectures to better capture temporal information in a stream of images has received little attention in deep RL, and our work fills this void. Perhaps the closest to our motivation is the work of \citet{amiranashvili2018motion} who explicitly use optical flow as an extra input to the RL policy. However, this approach requires additional information and supervision signal to train the flow estimator, which could be unavailable or inaccurate in practice. In contrast, our approach is a simple modification to existing deep RL architectures and does not require any additional auxiliary tasks or supervision signals.

\textbf{Two-Stream Video Classification} In video classification tasks, such as activity recognition~\citep{soomro2012ucf101}, there are a large body of works on how to utilize temporal information~\citep{donahue2015long, ji20123d, tran2015learning, carreira2017quo, wang2018non, feichtenhofer2019slowfast}. Of particular relevance is the two-stream architecture of \citet{simonyan2014two}, where one CNN stream takes the usual RGB frames, while the other the optical flow computed from the RGB values. The features from both streams are then late-fused to predict the activity class. That the two-stream architecture yields a significant performance gain compared to the single RGB stream counterpart, indicating the explicit temporal information carried by the flow plays an essential role in video understanding. Instead of directly computing the optical ﬂow, we propose to capture the motion information in latent space to avoid computational overheads and potential flow approximation errors. Our approach also could focus on domain-speciﬁc motions that might be overlooked in a generic optical ﬂow representation.

\section{Motivation}
\label{sec:motivation} 

\begin{figure}[h]
\begin{center}
\centering
\includegraphics[width=\linewidth]{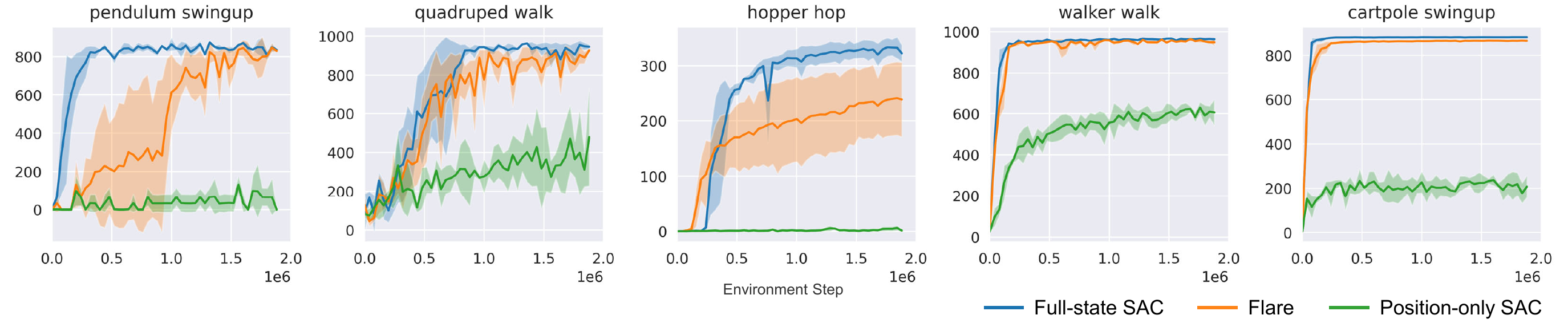}
\end{center}
\vspace{-3mm}
\small
\caption{Flare enables an RL agent with only access to positional state to recover a near-optimal policy relative to RL with access to the full state. In the above learning curves we show test-time performance for (i) full-state SAC (blue), where both pose and temporal information is given (ii) position-only SAC (green), and (iii) state-based Flare (orange), where only pose information is provided and velocities are approximated through pose offsets. Unlike full-state SAC, which learns the optimal policy, position-only SAC either fails or converges at suboptimal policies. Meanwhile, the fusion of positions and approximated velocities in Flare efficiently recovers near-optimal policies in most cases. This motivates using Flare for pixel-based input, where velocities are not present in the observation. These results show mean performance with standard deviations averaged over 3 seeds.}
\label{fig:state_sac}
\vspace{-3mm}
\end{figure}
We motivate Flare by investigating the importance of temporal information in state-based RL. Our investigation utilizes 5 diverse DMControl~\citep{tassa2018deepmind} tasks. The full state for these environments includes both the agent's pose information, such as the joints' positions and angles, as well as temporal information, such as the joints' translational and angular velocities. We train two variants with SAC\textemdash one where the agent receives the full state as input (full-state SAC), and the other with the temporal information masked out, i.e. the agent only receives the pose information as its input (position-only SAC). The resulting learning curves are in Figure~\ref{fig:state_sac}. While the full-state SAC learns the optimal policy quickly, the position-only SAC learns much sub-optimal policies, which often fail entirely.
Therefore, we conclude that effective policies cannot be learned from positions alone, and that temporal information is crucial for efficient learning.

While full-state SAC can receive velocity information from internal sensors in simulation, in the more general case such as learning from pixels, such information is often not readily available. For this reason, we attempt to approximate temporal information as the difference between two consecutive states' positions. Concretely, we compute the positional offset $\delta_{t} {=} (s^p_t {-} s^p_{t-1}, s^p_{t{-}1}{-}s^p_{t{-}2}, s^p_{t{-}2}{-}s^p_{t{-}3})$, and provide the fused vector $(s^p_t,\delta_t)$ to the SAC agent. This procedure describes the state-based version of Flare. Results shown in Figure~\ref{fig:state_sac} demonstrate that state-based Flare significantly outperforms the position-only SAC. Furthermore, it achieves optimal asymptotic performance and a learning efficiency comparable to full-state SAC in most environments. 

\begin{figure}[h]
\begin{center}
\centering
\includegraphics[width=\linewidth]{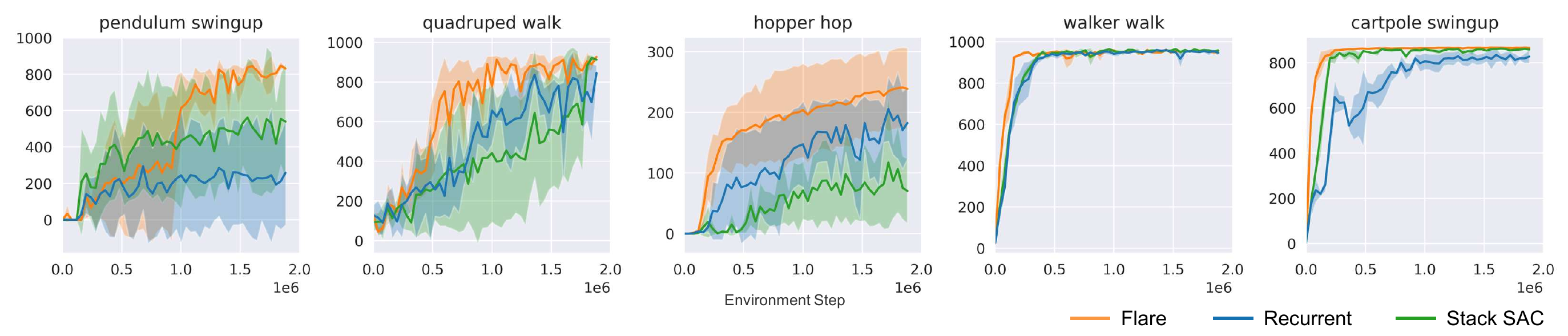}
\end{center}
\small
\vspace{-3mm}
\caption{We compare 3 ways to incorporate temporal information: i) Flare (orange) receives $(s^p_t, s^p_t {-} s^p_{t-1}, s^p_{t-1} {-} s^p_{t-2}, s^p_{t-2} {-} s^p_{t-3})$, 
ii) stack SAC (green) stacks $(s^p_t,s^p_{t{-}1}, s^p_{t{-}2}, s^p_{t{-}3})$ as inputs, and iii) recurrent SAC (blue) uses recurrent layers to process $(s^p_t,s^p_{t-1}, s^p_{t-2}, s^p_{t-3})$. Stack SAC and recurrent SAC perform significantly worse than Flare on most environments, highlighting the benefit of how Flare handles temporal information. Results are averaged over 3 seeds.
}\label{fig:state_ablation}
\vspace{-3mm}
\end{figure}

Given that the position-only SAC utilizes $s^p_t$ compared to Flare that utilizes $s^p_t$ and $\delta_{t}$, we also investigate a variant (stack SAC) where the SAC agent takes consecutive positions 
$(s^p_t,s^p_{t-1}, s^p_{t-2}, s^p_{t-3})$. Stack SAC reflects the frame-stack heuristic used in pixel-based RL. Results in Figure~\ref{fig:state_ablation} show that Flare still significantly outperforms stack SAC. It suggests that the well-structured inductive bias in the form of temporal-position fusion is essential for efficient learning.

Lastly, since a recurrent structure is an alternative approach to process temporal information, we implement an SAC variant with recurrent modules (Recurrent SAC) to compare with Flare. Specifically, we pass a sequence of poses $s^p_t, s^p_{t-1}, s^p_{t-2}, s^p_{t-3}$ through an LSTM cell. The number of the LSTM hidden units $h$ is set to be the same as the dimension of $\delta_t$ in Flare. 
The trainable parameters of the LSTM cell are updated to minimize the critic loss. Recurrent SAC is more complex to implement and requires longer wall-clock training time, but performs worse than Flare as shown in Figure~\ref{fig:state_ablation}.

Our findings from the state experiments in Figure~\ref{fig:state_sac} and Figure~\ref{fig:state_ablation} suggest that (i) temporal information is crucial to learning effective policies in RL,  (ii) using Flare to approximate temporal information in the absence of sensors that provide explicit measurements is sufficient in most cases, and (iii) to incorporate temporal information via naively staking position states or a recurrent module are less effective than Flare. In the next section, we carry over these insights to pixel-space RL.    
\begin{figure}[t]
\begin{center}
\centering
\includegraphics[width=\linewidth]{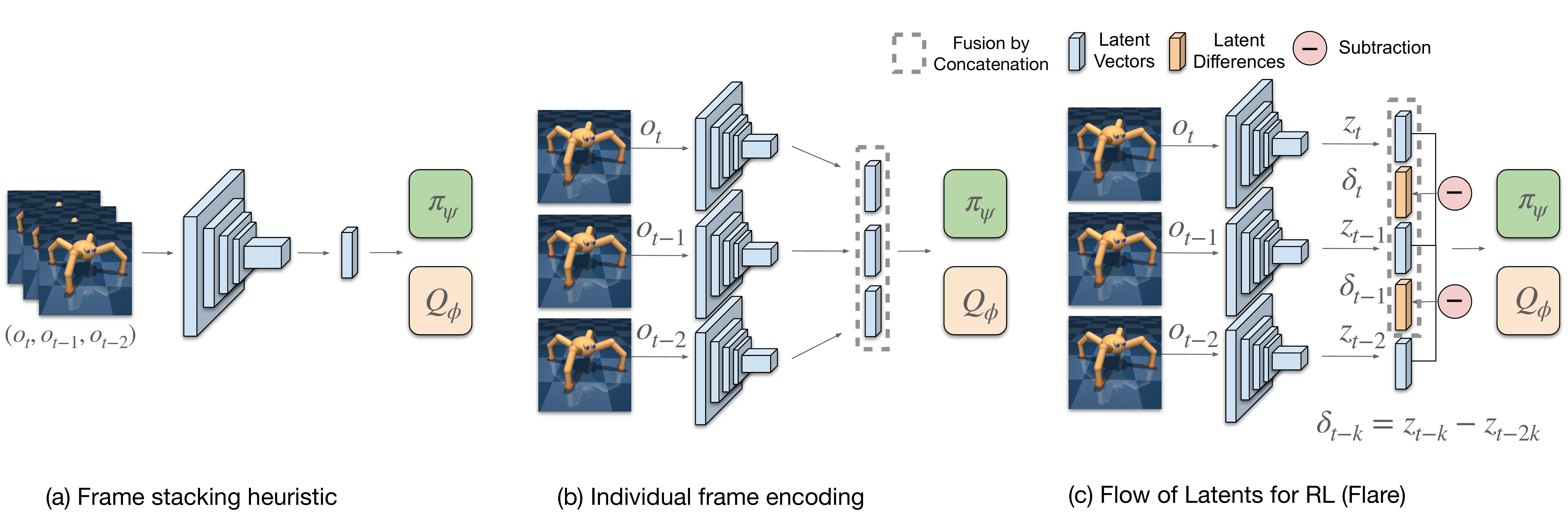}
\end{center}
\caption{{\bf F}low of {\bf La}tents for {\bf Re}inforcement Learning (\emph{Flare}): (a) the architecture for the frame stacking heuristic, (b) an alternative to the frame stacking hueristic by encoding each image individually, and (c) the Flare architecture which encodes images individually, computes the feature differences, and fuses the differences together with the latents.}\label{fig:flare}
\end{figure}
\section{Reinforcement Learning with Latent Flow}
To date, frame stacking is the most common way of pre-processing pixel-based input to convey temporal information for RL algorithms. This heuristic, introduced by \citet{mnih2015human}, has been largely untouched since its inception and is used in most state-of-the-art RL architectures. However, our observations from the experiments run on state inputs in Section~\ref{sec:motivation} suggest an alternative to the frame stacking heuristic through the explicit inclusion of temporal information as part of the input. Following this insight, we seek a general alternative approach to explicitly incorporate temporal information that can be coupled to any base RL algorithm with minimal modification. To this end, we propose the {\bf F}low of {\bf La}tents for {\bf Re}inforcement Learning (\emph{Flare}) architecture. Our proposed method calculates differences between the latent encodings of individual frames and fuses the feature differences and latent embeddings before passing them as input to the base RL algorithm, as shown in Figure~\ref{fig:flare}. We demonstrate Flare on top of 2 state-of-the-art model-free off-policy RL baselines, RAD-SAC~\citep{laskin_lee2020rad} and Rainbow DQN~\citep{hessel2017rainbow}, though in principle any RL algorithm can be used in principle.

\subsection{Latent Flow}\label{sec:lf}
In computer vision, the most common way to explicitly inject temporal information of a video sequence is to compute dense optical flow between consecutive frames~\citep{simonyan2014two}. Then the RGB and the optical flow inputs are individually fed into two streams of encoders and the features from both are fused in the later stage of the piple. But two-stream architectures with optical flow are not as applicable to RL, because it is too computationally costly to generate optical flow on the fly.

%
%
To address this challenge and motivated by experiments in Section~\ref{sec:motivation}, we propose an alternative architecture that is similar in spirit to the two-stream networks for video classification. Rather than computing optical flow directly, we approximate temporal information in the latent space. Instead of encoding a stack of frames at once, we use a frame-wise CNN to encode each individual frame. Then we compute the differences between the latent encodings of consecutive frames, which we refer to as \emph{latent flow}. Finally, the latent features and the latent flow are fused together through concatenation before getting passed to the downstream RL algorithm. We call the proposed architecture as {\bf F}low of {\bf La}tents for {\bf Re}inforcement Learning (\emph{Flare}). 

\subsection{Implementation Details}\label{sec:archit}
\begin{wrapfigure}[13]{L}{0.5\textwidth} 
  \begin{algorithm}[H]                
    
    \SetCustomAlgoRuledWidth{0.45\textwidth}  
    \caption{Pixel-based Flare Inference}\label{algo:flare}
       Given $\pi_\psi$, $f_\mathrm{CNN}$\;
       \For{each environment step $t$}
       {
       $z_j{=}f_{\mathrm{CNN}}(o_j), j{=} t{-}k,..,t$\;
       $\delta_j{=}z_{j} {-} z_{j-1}, j{=}t{-}k{+}1,..,t$\;
       $\mathbf{z}_t{=}(z_{t{-}k{+}1},{\cdots} , z_t, \delta_{t{-}k{+}1}, {\cdots} , \delta_{t})$\;
       $\mathbf{z_t} = \mathrm{LayerNorm} (f_{\mathrm{FC}}(\mathbf{z_t}))$\;
       $a_t{\sim}\pi_\psi(a_t|\mathbf{z}_t)$\;
       $o_{t+1}{\sim}p(o_{t{+}1}|a_t, \mathbf{o}_t = (o_t, o_{t{-}1}..o_{t{-}k}))$\;
       }
  \end{algorithm}

\end{wrapfigure}
For clarity of exposition, we select RAD as the base algorithm to elaborate the execution of Flare. Also, we use RAD later on in our experiments as the comparative baseline (Section~\ref{sec:exp}). The RAD architecture, shown in Figure~\ref{fig:flare}a, stacks multiple data augmented frames observed in the pixel space and encodes them altogether through an CNN. This can be viewed as a form of early fusion~\citep{KarpathyVideo}. 

Another preprocessing option is to encode each frame individually through a shared frame-wise encoder and perform late fusion of the resulting latent features, as shown in Figure~\ref{fig:flare}b. However, we find that simply concatenating the latent features results in inferior performance when compared to the frame stacking heuristic, which we further elaborate in Section~\ref{sec:ablate}. We conjecture that pixel-level frame stacking benefits from leveraging both the CNN and the fully connected layers to process temporal information, whereas latent-level stacking does not propagate temporal information back through the CNN encoder. 

Based on this conjecture, we explicitly compute the latent flow $\delta_t = z_{t} - z_{t-1}$ while detaching the $z_{t-1}$ gradients when computing $\delta_t$. We then fuse together $(\delta_t, z_t)$. Next, since negative values in the fused latent embedding now possesses semantic meaning from $\delta_t$, instead of ReLU non-linearity, we pass the embedding through a fully-connected layer followed by layer normalization, before entering the actor and critic networks as shown in Figure~\ref{fig:flare}c. Pseudocode illustrates inference with Flare in Algorithm~\ref{algo:flare}; during training, the encodings of latent features and flow are done in the same way except with augmented observations. 
\begin{table}
\centering
\begin{tabular}{c|cc|cc}
Task & Flare (500K) & RAD (500K) & Flare (1M) & RAD (1M)\\ \hline
Quadruped Walk & $\mathbf{296}\pm139$ & $206\pm112$ & $\mathbf{488}\pm221$ & $322\pm229$ \\
Pendulum Swingup & $\mathbf{242}\pm152$ & $79\pm73$ & $\mathbf{809}\pm31$ & $520\pm321$\\ 
Hopper Hop & $\mathbf{90}\pm55$ & $40\pm41$ & $\mathbf{217}\pm59$ & $211\pm27$ \\
Finger Turn hard & $\mathbf{282}\pm67$ & $137\pm98$ & $\mathbf{661}\pm315$ & $249\pm98$ \\
Walker Run & $426\pm33$ & $\mathbf{547}\pm48$ & $556\pm93$ & $\mathbf{628}\pm39$ \\
\end{tabular}
\vspace{-2mm}
\caption{\small{Evaluation on  5  benchmark  tasks around  500K and 1M environment steps. We evaluate over 5 seeds, each of 10 trajectories and show the mean $\pm$ standard deviation across runs. 
}}\label{tab:results}
\vspace{-3mm}
\end{table}
\begin{table}
\centering
\begin{tabular}{ccc|ccc}
 & Rainbow & Flare & & Rainbow & Flare \\ \hline
Assault & $\mathbf{15229}{\pm}3603$ & $12724{\pm}1107$ & Breakout & $280{\pm}18$ & $\mathbf{345}{\pm}22$ \\
Berserk & $1636{\pm}598$ & $\mathbf{2049}{\pm}421$ & Defender & $44694{\pm}3984$ & $\mathbf{86982}{\pm}29214$\\ 
Montezuma & $900{\pm}807$ & $\mathbf{1668}{\pm}1055$ & Seaquest & $\mathbf{24090}{\pm}12474$ & $13901{\pm}8085$ \\
Phoenix & $16992{\pm}3295$ & $\mathbf{60974}{\pm}18044$ & Tutankham & $\mathbf{247}{\pm}11$& $\mathbf{248}{\pm}20$ \\
\end{tabular}
\vspace{-2mm}
\caption{\small{Evaluation on  8 benchmark  Atari games at 100M training steps over 5 seeds. 
}}\label{tab:atari}
\vspace{-6mm}
\end{table}
\vspace{-3mm}
\section{Experiments}\label{sec:exp}
\vspace{-1mm}
\begin{figure}[t]
\begin{center}
\includegraphics[width=0.99\linewidth]{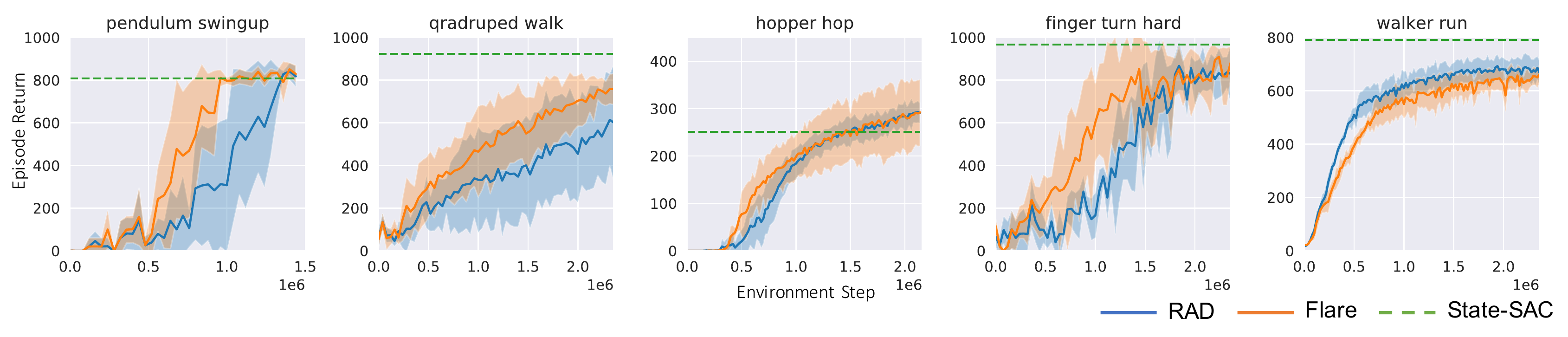}
\end{center}
\small
\caption{We compare Flare and the current STOA model-free baseline RAD on 5 challenging DMControl environments. Pendulum Swingup are trained over $1.5\mathrm{e}6$ and the rest $2.5\mathrm{e}6$. Flare substantially outperforms RAD on a majority (3 out of the 5) of environments, while being competitive in the remaining. 
Results are averaged over 5 random seeds with standard deviation (shaded regions). 
}
\label{fig:core}
\end{figure}
\begin{figure}[h]
\begin{center}
\includegraphics[width=0.99\linewidth]{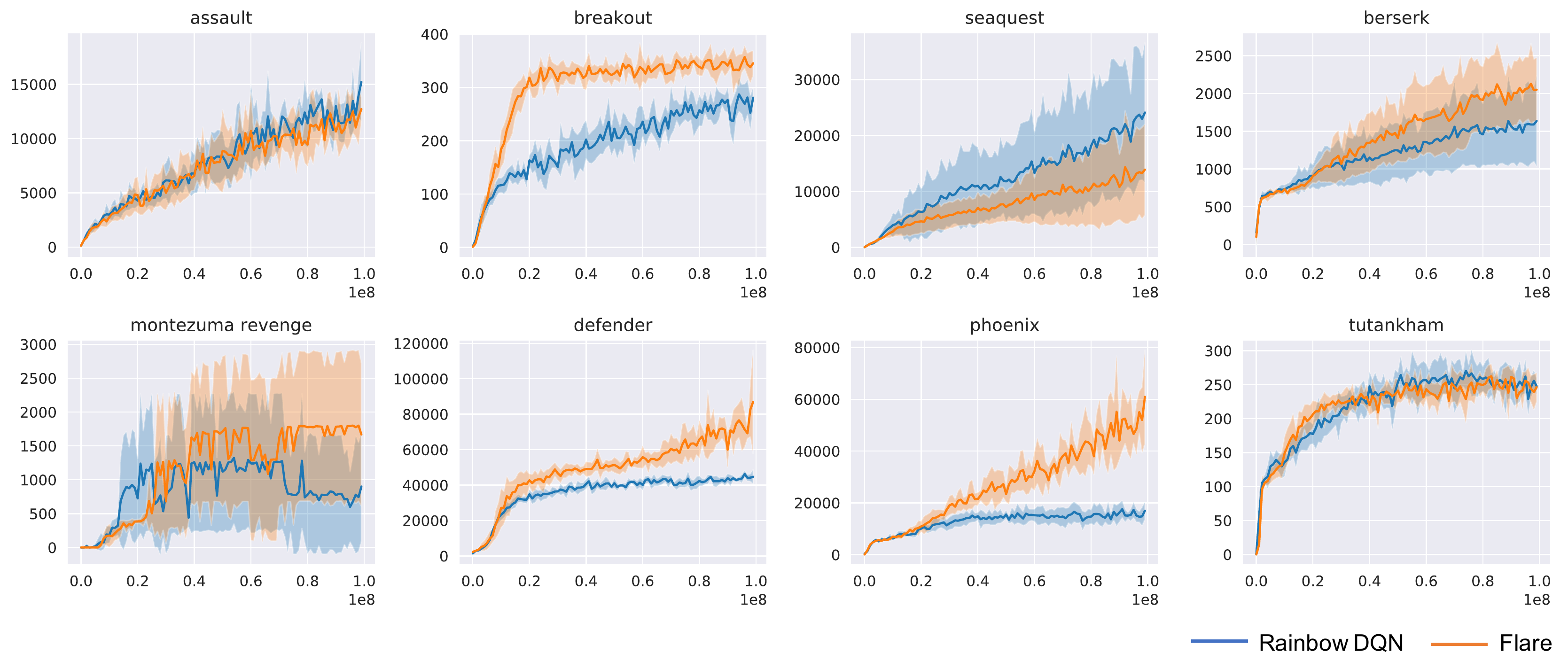}
\end{center}
\small
\caption{We compare Rainbow DQN and Flare on 8 Atari games over 100M training steps. Flare substantially enhances 5 out of 8 games over the baseline Rainbow DQN while matching the rest except Seaquest. Results are averaged over 5 random seeds with standard deviation (shaded regions).
}
\label{fig:atari}
\end{figure}

\begin{figure}[h]
\begin{center}
\includegraphics[width=0.99\linewidth]{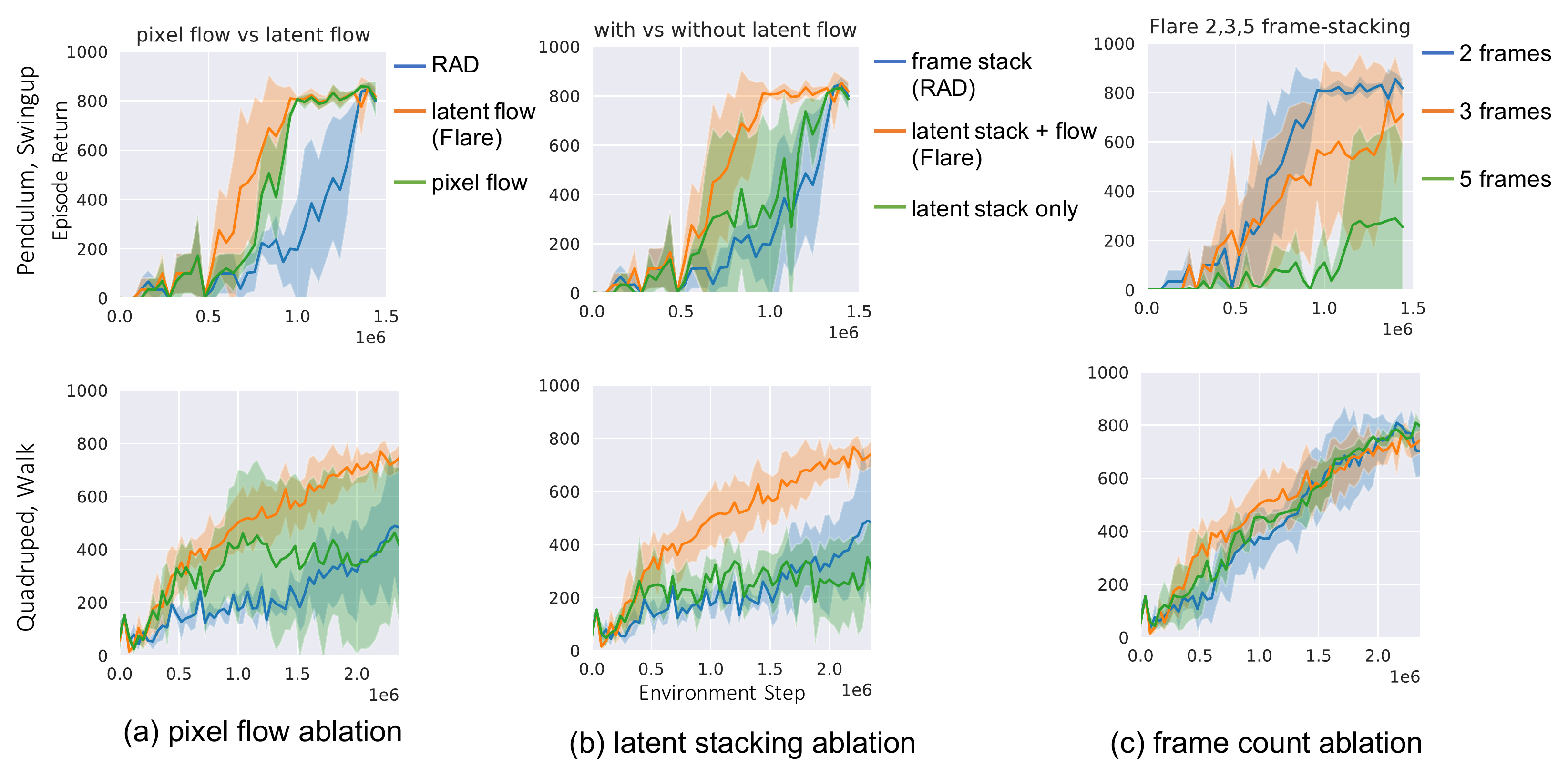}
\end{center}
\small
\caption{We perform 3 ablation studies: (a) {\it Pixel flow ablation}: we compare using pixel-level and latent-level (Flare) differences. Flare is more stable and performs better. (b) {\it Latent stack ablation}: we compare using latent stack with and without the latent flow. The latter performs significantly worse, suggesting that the latent flow is crucial. (c) {\it Frames count ablation}: we test using different number of frames for Flare.}\label{fig:ablate}
\vspace{-5mm}
\end{figure}
%

In this section, we first present the main experimental results, where we show that Flare achieves substantial performance gains over the base algorithm RAD~\citep{laskin_lee2020rad}. Then we conduct a series of ablation studies to stress test the design choices of the Flare architecture.
In the appendix, we introduce the 5 continuous control tasks from DMControl suite~\citep{tassa2018deepmind} and 8 Atari games~\citep{bellemare2013arcade} that our experiments focus on in the Appendix.
\vspace{-3mm}
\subsection{Main Results}\label{sec:results}
\textbf{DMControl:} Our main experimental results on the 5 DMControl tasks are presented in Figure~\ref{fig:core} and Table~\ref{tab:results}. We find that Flare outperforms RAD in terms of both final performance and sample efficiency for majority (3 out of 5) of the environments, while being competitive on the remaining environments. Specifically, Flare attains similar asymptotic performance to state-based RL on Pendulum Swingup, Hopper Hop, and Finger Turn-hard. For Quadruped Walk, a particularly challenging environment due to its large action space and partial observability, Flare learns much more efficiently than RAD and achieves a higher final score. Moreover, Flare outperforms RAD in terms of sample efficiency on all of the core tasks except for Walker Run as shown in Figure~\ref{fig:core}. The 500k and 1M environment step evaluations in Table~\ref{tab:results} show that, on average, Flare achieves $\textbf{1.9} \times$  and $\textbf{1.5} \times$ higher scores than RAD at the 500k step and the 1M step benchmarks, respectively. 
%

\textbf{Atari:} The results on the 8 Atari games are in Figure~\ref{fig:atari} and Table~\ref{tab:atari}. Here the baseline Rainbow DQN's model architecture is modified to match that of Flare, including increasing the number of last layer convolutional channels and adding a fully-connected layer plus layer normalization before the Q networks. Again, we observe substantial performance gain from Flare on the majority (5 out of 8) of the games, including the challenging Montezuma's Revenge. On most of the remaining games, Flare is equally competitive except for Seaquest.
In Appendix~\ref{sec:atari}, we also show that Flare performs competitively when comparing against other DQN variants at 100M training steps, including the original Rainbow implementations.
\subsection{Ablation Studies}\label{sec:ablate}
We ablate a number of components of the Flare architecture on the Quadruped Walk and Pendulum Swingup environments to stress test the Flare architecture. The results shown in Figure~\ref{fig:ablate} aim to answer the following questions:

{\bf  Q1}: {\it Do we need latent flow or is computing pixel differences sufficient?} \\
{\bf A1}: Flare proposes a late fusion of latent differences with the latent embeddings, while a simpler approach is an early fusion of pixel differences with the pixel input, which we call pixel flow. We compare Flare to pixel flow in Figure~\ref{fig:ablate} (left) and find that pixel flow is above RAD but significantly less efficient and less stable than Flare, particularly on Quadruped Walk. This ablation suggests that late fusion temporal information after encoding the image is preferred to early fusion. 


{\bf  Q2}: {\it Are the gains coming from latent flow or individual frame-wise encoding?}\\ 
{\bf A2}: Next, we address the potential concern that the performance gain of Flare stems from the frame-wise ConvNet architectural modification instead of the fusion of latent flow. Concretely, we follow the exact architecture and training as Flare, but instead of concatenating the latent flow, we concatenate each frame's latent vector after the convolution encoders directly as described in Figure~\ref{fig:flare}b. This ablation is similar in spirit to the state-based experiments in Figure~\ref{fig:state_ablation}. The learning curves in Figure~\ref{fig:ablate} (center) show that individual frame-wise encoding is not the source of the performance lift: frame-wise encoding, though on par with RAD on Pendulum Swingup, performs significantly worse on Quadruped Walk. Flare's improvements over RAD are therefore most likely a result of the explicit fusion of latent flow.

{\bf  Q3}: {\it How does the input frame count affect performance?} \\
{\bf A3}: Lastly, we compare stacking 2, 3, and 5 frames in Flare in Figure~\ref{fig:ablate} (right). We find that changing the number of stacked frames does not significantly impact the locomotion task, quadruped walk, but Pendulum Swingup tends to be more sensitive to this hyperparameter. Interestingly, the optimal number of frames for Pendulum Swingup is 2, and more frames can in fact degrade Flare's performance, indicating that the immediate position and velocity information is the most critical to learn effective policies on this task. We hypothesize that Flare trains more slowly with increased frame count on Pendulum Swingup due to the presence of unnecessary information that the actor and critic networks need to learn to ignore. 

\section{Conclusion}
We propose Flare, an architecture for RL that explicitly encode temporal information by computing flow in the latent space. In experiments, we show that in the state space, Flare can recover the optimal performance with only state positions and no access to the state velocities. In the pixel space, Flare improves upon the state-of-the-art model-free RL algorithms on the majority of selected tasks in the DMControl and Atari suites, while matching in the remaining. 
\bibliography{main}
\bibliographystyle{iclr2021_conference}
\appendix
\newpage
\section{Appendix}
\subsection{Background}\label{sec:background}

    \textbf{Soft Actor Critic} (SAC)~\citep{haarnoja2018soft} is an off-policy actor-critic RL algorithm for continuous control with an entropy maximization term augmented to its score function to encourage exploration. SAC learns a policy network $\pi_{\psi}(a_t|\mathbf{o}_t)$ and critic networks $Q_{\phi_1}(\mathbf{o}_t, a_t)$ and $Q_{\phi_2}(\mathbf{o}_t, a_t)$ to estimate state-action values. 
    The critic $Q_{\phi_i}(\mathbf{o}_t, a_t)$ is optimized to minimize the (soft) Bellman residual error:
    \begin{equation}
    \label{eq:sac_critic_loss}
    \mathcal{L}_Q(\phi_i) = \E_{\tau\sim \mathcal{B}}\:\left[\big(Q_{\phi_i}(\mathbf{o}_t, a_t) - (r_t + \gamma V(\mathbf{o}_{t+1}))\big)^2\right],
    \end{equation}
    where $r$ is the reward, $\gamma$ the discount factor, $\tau = (\mathbf{o}_t, a_t, \mathbf{o}_{t+1}, r_t)$ is a transition sampled from replay buffer $\mathcal{B}$, and $V(\mathbf{o}_{t+1})$ is the (soft) target value estimated by:
    \begin{equation}
    \label{eq:sac_value}
    V(\mathbf{o}_{t+1})=\left(\min_{i} Q_{\bar \phi_i}(\mathbf{o}_{t+1}, a_{t+1}) - \alpha\,\mathrm{log}\,\pi_{\psi}(a_{t+1}|\mathbf{o}_{t+1})]\right),
    \end{equation}
    where $\alpha$ is the entropy maximization coefficient. 
    For stability, in eq.~\ref{eq:sac_value}, $Q_{\bar \phi_i}$ is the exponential moving average of $Q_{\phi_i}$'s over training iterations.  
    The policy $\pi_{\psi}$ is trained to maximize the expected return estimated by $Q$ together with the entropy term
    \begin{equation}
    \label{eq:sac_policy_loss}
    L_\pi(\psi)=-\E_{a_t\sim\pi}\:[\min_{i}Q_{\phi_i}(\mathbf{o}_{t}, a_t) - \alpha \,\mathrm{log}\,\pi_\psi(a_t|\mathbf{o}_t)],
    \end{equation}
    where $\alpha$ is also a learnable parameter. 
    
\textbf{Reinforcement Learning with Augmented Data} (RAD)~\citep{laskin_lee2020rad} is a recently proposed training technique. In short, RAD pre-processes raw pixel observations by applying random data augmentations, such as random translation and cropping, for RL training.
As simple as it is, RAD has taken many existing RL algorithms, including SAC, to the next level. 
For example, on many DMControl~\citep{tassa2018deepmind} benchmarks, while vanilla pixel-based SAC performs poorly, RAD-SAC\textemdash i.e. applying data augmentation to pixel-based SAC\textemdash achieves state-of-the-art results both in sample efficiency and final performance. 
In this work, we refer RAD to RAD-SAC and the augmentation used is random translation. 
    
\textbf{Rainbow DQN} is an extension of the Nature Deep Q Network (DQN)~\citep{mnih2015human}, which combines multiple follow-up improvements of DQN to a single algorithm~\citep{hessel2017rainbow}. In summary, DQN~\citep{mnih2015human} is an off-policy RL algorithm that leverages deep neural networks (DNN) to estimate the Q value directly from the pixel space. The follow-up works Rainbow DQN bring together to enhance the original DQN include double Q learning~\citep{hasselt2010double}, prioritized experience replay~\citep{schaul2015prioritized}, dueling network~\citep{wang2016dueling}, noisy network~\citep{fortunato2017noisy}, distributional RL~\citep{bellemare2017distributional} and multi-step returns~\citep{SuttonBook}. Rainbow DQN is one of the state-of-the-art RL algorithms on the Atari 2600 benchmark~\citep{bellemare2013arcade}. We thus adopt an official implementation of Rainbow~\citep{dqnzoo2020github} as our baseline to directly augment Flare on top.
\subsection{Environments and Evaluation Metrics}\label{sec:dmcontrol}
\begin{figure}[t]
\begin{center}
\includegraphics[width=0.95\linewidth]{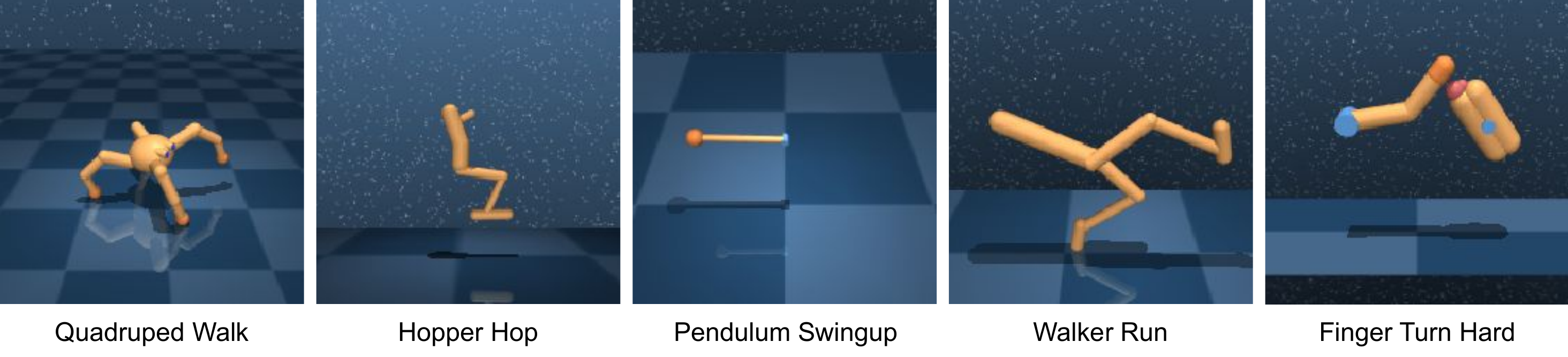}
\end{center}
\small
\caption{We choose the following environments for our main experiments -- (i) quadruped walk, which requires coordination of multiple joints, (ii) hopper hop, which requires hopping while maintaining balance, (iii) pendulum swingup, an environment with sparse rewards, (iv) walker run, which requires the agent to maintain balance at high speeds, and (v) finger turn hard, which requires precise manipulation of a rotating object. These environments are deemed challenging because prior state-of-the-art model-free pixel-based methods \citep{laskin_srinivas2020curl,kostrikov2020image,laskin_lee2020rad} either fail to reach the asymptotic performance of state SAC or learn less efficiently.  }
\label{fig:envs}
\vspace{-3mm}
\end{figure}

    \textbf{The DeepMind Control Suite} (DMControl)~\citep{tassa2018deepmind}, based on MuJoCo~\citep{mujoco}, is a commonly used benchmark for continuous control from pixels.
    Prior works such as DrQ~\citep{kostrikov2020image} and RAD~\citep{laskin_lee2020rad} have made substantial progress on this benchmark and closed the gap between state-based and pixel-based efficiency on the simpler environments in the suite, such as Reacher Easy, Ball-in-cup Catch, Finger Spin, Walker Walk, Cheetah Run, Cartpole Swingup. However, current pixel-based RL algorithms struggle to learn optimal policies efficiently in more challenging environments that feature partial observability, sparse rewards, or precise manipulation. In this work, we study more challenging tasks from the suite to better showcase the efficacy of our proposed method. The 5 environments, listed in Figure~\ref{fig:envs}, include Walker Run (requires maintaining balance with speed), Quadruped Walk (partially observable agent morphology), Hopper Hop (locomotion with sparse rewards), Finger Turn-hard (precise manipulation), and Pendulum Swingup (torque control with sparse rewards). For evaluation, we benchmark performance at 500K and 1M \emph{environment steps} and compare against RAD.
    
    \textbf{The Atari 2600 Games}~\citep{bellemare2013arcade} is another highly popular RL benchmark. Recent efforts have let to a range of highly successful algorithms~\citep{espeholt2018impala, hessel2017rainbow, kapturowski2018recurrent, hafner2019dream, agent57} to solve Atari games directly from pixel space. A representative state-of-the-art is Rainbow DQN (see Section~\ref{sec:background}). We adopt the official Rainbow DQN implementation~\citep{dqnzoo2020github} as our baseline. Then we simply modify the model architecture to incorporate Flare while retaining all the other default settings, including hyperparameters and preprocessing. To ensure comparable model capacity, the Flare network halves the number of convolutional channels and adds a bottleneck FC layer to reduce latent dimension before entering the Q head (code in the Supplementary Materials). We evaluate on a diverse subset of Atari games at 50M \emph{training steps}, namely Assault, Breakout, Freeway, Krull, Montezuma Revenge, Seaquest, Up n Down and Tutankham, to assess the effectiveness of Flare.

\subsection{Compare Flare with DQN Variants on Atari}
\label{sec:atari}
\begin{table}[h]
\centering
\smaller 
\begin{tabular}{c|cccccccc}
 &Flare&Rainbow& Rainbow$^\dagger$& Prioritized$^\dagger$ & QR DQN$^\dagger$ & IQN$^\dagger$ \\ \hline
Assault & $12724{\pm}1107$ & $\mathbf{15229}{\pm}2061$ &  $13194{\pm}4348$ & $10163{\pm}1558$ & $10215{\pm}1255$ & $11903{\pm}1251$  \\
Breakout & $345{\pm}22$ & $280{\pm}18$ &  $366{\pm}20$ & $359{\pm}33$ &  $450{\pm}29$ & $\mathbf{492}{\pm}86$ \\
Berzerk & $\mathbf{2049}{\pm}421$ & $1636{\pm}598$ &  $3151{\pm}537$ & $986{\pm}98$ & $896{\pm}98$ & $946{\pm}48$  \\
Defender & $\mathbf{86982}{\pm}29214$ & $44694{\pm}3984$ &  $52419{\pm}4481$ & $13750{\pm}4182$ & $32320{\pm}8997$ & $33767{\pm}3643$  \\ 
Montezuma & $\mathbf{1668}{\pm}1055$ & $900{\pm}807$ &  $80{\pm}160$ & $0{\pm}0$  & $0{\pm}0$ & $0{\pm}0$\\
Seaquest & $13901{\pm}8085$ & $\mathbf{24090}{\pm}12474$ &  $5838{\pm}2450$ & $7436{\pm}1790$  & $14864{\pm}3625$ & $16866{\pm}4539$  \\
Phoenix & $60974{\pm}18044$ & $16992{\pm}3295$ & $\mathbf{82234}{\pm}33388$ & $10667{\pm}3142$  & $41866{\pm}3673$ & $35586{\pm}3510$ \\
Tutankham & $\mathbf{248}{\pm}20$& $\mathbf{247}{\pm}11$ & $214{\pm}24$ & $168{\pm}22$  & $171{\pm}30$  & $216{\pm}34$\\
\end{tabular}
\caption{\small{Evaluation on  8 benchmark  Atari games at 100M training steps over 5 seeds. $\dagger$ directly taken from DQN Zoo repository and the rest are collected from our experiments.  
}}\label{tab:atari}
\end{table}

\subsection{Implementation Details for Pixel-based Experiments}

\begin{center}
\begin{tabular}{l l}
 \hline
 \textbf{Hyperparameter} & \textbf{Value} \\ [0.5ex] 
 \hline
 Augmentation & Random Translate   \\ 
 Observation size & $(100, 100)$   \\  
 Augmented size & $(108, 108)$   \\
 Replay buffer size & $100000$ \\
 Initial steps & $10000$ \\
 Training environment steps & $1.5e6$ pendulum swingup\\
 &$2.5e6$ others\\
 Batch size & $128$ \\
 Stacked frames & $2$ pendulum swingup \\
                & $3$ others \\
 Action repeat & $2$ walker run, hopper hop \\
                & $4$ others \\
 Camera id & $2$ quadruped, walk \\
            & $0$ others \\
 Evaluation episode length & $10$ \\
 Hidden units (MLP) & $1024$ \\
 Number of layers (MLP) & $2$ \\ 
 Optimizer & Adam \\
 $(\beta_1, \beta_2)\rightarrow(f_{CNN}, \pi_\psi, Q_\phi)$ & $(.9, .999)$ \\
 $(\beta_1, \beta_2) \rightarrow (\alpha)$ & $(.5, .999)$ \\
 Learning Rate $(\pi_\psi, Q_\phi)$ & $2e-4$ \\
 Learning Rate $(f_{CNN})$ & $1e-3$ \\
 Learning Rate $(\alpha)$ & $1e-4$ \\ 
 Critic target update frequency & $2$\\
 Critic EMA $\tau$ & $0.01$ \\
 Encoder EMA $\tau$ & $0.05$ \\
 Convolutional layers & $4$ \\
 Number of CNN filters & $32$ \\
 Latent dimension & $64$ \\
 Non-linearity & ReLU \\
 Discount $\gamma$ & $0.99$ \\
 Initial Temperature & $0.1$ \\
 \hline
\end{tabular}
\end{center}

\subsection{Implementation Details for State-based Motivation Experiments}

\begin{center}
\begin{tabular}{l l}
 \hline
 \textbf{Hyperparameter} & \textbf{Value} \\ [0.5ex] 
 \hline
 Replay buffer size & $2000000$ \\
 Initial steps & $5000$ \\
 Batch size & $1024$ \\
 Stacked frames & $4$ Flare, Stack SAC;  \\
                & $1$ otherwise \\
 Action repeat & $1$ \\
 Evaluation episode length & $10$ \\
 Hidden units & $1024$ \\
 Number of layers & $2$ \\ 
 Optimizer & Adam \\
 $(\beta_1, \beta_2)\rightarrow(f_{CNN}, \pi_\psi, Q_\phi)$ & $(.9, .999)$ \\
 $(\beta_1, \beta_2) \rightarrow (\alpha)$ & $(.9, .999)$ \\
 Learning Rate $(\pi_\psi, Q_\phi)$ & $1e-4$ \\
 Learning Rate $(\alpha)$ & $1e-4$ \\ 
 Critic target update frequency & $2$\\
 Critic EMA $\tau$ & $0.01$ \\
 Non-linearity & ReLU \\
 Discount $\gamma$ & $0.99$ \\
 Initial Temperature & $0.1$ \\
 \hline
\end{tabular}
\end{center}

\subsection{Interpreting FLARE from a Two-stream Prospective} \label{sec: two-stream_flare}
Let $f_{\mathrm{CNN}}$ and $o_t'$  be the pixel encoder and the augmented observation in Flare. Then, $z_t = f_{\mathrm{CNN}}(o_t')$ denotes the latent encoding for a frame at time $t$. By computing the latent flow $\delta_t = z_{t} - z_{t-1}$, Flare essentially approximates $\frac{\partial z_t}{\partial t}$ via backward finite difference. Then following chain rule, we have
$$\frac{\partial z}{\partial t} = \frac{\partial z}{\partial o'} \cdot \frac{\partial o'}{\partial t} = \frac{f_{\mathrm{CNN}}(o')}{\partial o'} \cdot \mathrm{dense}\, \mathrm{optical}\, \mathrm{flow}$$
indicating that Flare eventually uses dense optical flow by propagating it through the derivative of the trained encoder. While the two-stream architecture trains a spatial stream CNN from RGB channels and a temporal stream from optical flow separately, Flare can be interpreted as training one spatial stream encoder from the RL objective and approximate the temporal stream encoder with its derivative.

\end{document}